\documentclass[twocolumn]{article}

\usepackage{ifluatex,ifxetex}
\ifluatex
  \usepackage{fontspec} % optional
\else\ifxetex
  \usepackage{fontspec} % optional
\else % assume that pdftex engine is in use
  \usepackage[T1]{fontenc}
  \usepackage[utf8]{inputenc} % optional for "\ng"
\fi\fi

\usepackage{array}
\usepackage{amsmath}
\usepackage{amssymb}
\usepackage{amsthm}
\usepackage{algorithm}
\usepackage{algorithmic}
\usepackage{authblk}

\usepackage{bbm}
\usepackage{booktabs}

\usepackage{caption}
\usepackage{cite}
\usepackage{csvsimple}
\usepackage{color}
\usepackage{csvsimple}

\usepackage{diagbox}
\usepackage{dsfont}

\usepackage{graphicx}
\usepackage{geometry}

\usepackage[pagebackref=true,breaklinks=true,letterpaper=true,colorlinks,bookmarks=false]{hyperref}

\usepackage{multirow}
\usepackage{makecell}
\usepackage{multirow}
\usepackage{makecell}
\usepackage{multirow}
\usepackage{makecell}

\usepackage{paralist}
\usepackage{pgfplots}

\usepackage{subfigure}

\usepackage{tikz}

\usepackage{url}

\usepackage{xspace}

\tikzset{mycolor/.code=\xdef\mycolor{#1},my color/.style={mycolor=#1,color=#1},
my pattern color/.style={mycolor=#1,pattern color=#1}}

\usetikzlibrary{pgfplots.groupplots}

\catcode`\_=11\relax

\def\_email#1@#2\q_nil{%
    \href{mailto:#1@#2}{{\emailfont #1\emailampersat #2}}
}
\newcommand\emailfont{\sffamily}
\newcommand\emailampersat{{\color{black}\small@}}
\catcode`\_=8\relax 

\newcommand{\PreserveBackslash}[1]{\let\temp=\\#1\let\\=\temp}
\newcolumntype{C}[1]{>{\PreserveBackslash\centering}p{#1}}

\newcommand{\modelname}{\mbox{MCH}\xspace}

\newcommand{\nuswide}{\mbox{NUS-WIDE}\xspace}
\newcommand{\coco}{\mbox{MS-COCO}\xspace}
\newcommand{\flickr}{\mbox{MIR FLICKR}\xspace}

\newcommand{\etoe}{\mbox{end-to-end}\xspace}

\def\a{{\bf a}}

\def\b{{\bf b}}

\def\h{{\bf h}}

\def\X{{\bf X}}

\def\y{{\bf y}}

\def\S{{\bf S}}
\def\x{{\bf x}}
\def\y{{\bf y}}
\def\z{{\bf z}}

\def\q{{\bf q}}

\def\u{{\bf u}}

\def\W{{\bf W}}
\def\w{{\bf w}}
\def\X{{\bf X}}
\def\z{{\bf z}}

\def\0{{\bf 0}}
\def\1{{\bf 1}}

\def\BM{{\mathcal B}}

\def\OM{{\mathcal O}}

\def\sgn{\text{sign}}

\def\1{\mathds{1}}

\newgeometry{left=2cm,right=2cm,top=2cm,bottom=2cm}

\newcolumntype{C}[1]{>{\centering\let\newline\\\arraybackslash\hspace{0pt}}m{#1}}
\newcolumntype{L}[1]{>{\let\newline\\\arraybackslash\hspace{0pt}}m{#1}}

\font\emailfont=cmr12 at 9pt

\begin{document}
\title{\bf Multiple Code Hashing for Efficient Image Retrieval}
\date{\today}
\author[ ]{Ming-Wei Li}
\author[ ]{Qing-Yuan Jiang}
\author[ ]{Wu-Jun Li}

\affil[ ]{National Key Laboratory for Novel Software Technology \protect\\
Collaborative Innovation Center of Novel Software Technology and Industrialization \protect\\
Department of Computer Science and Technology, Nanjing University, China}
\affil[ ]{\texttt{limw@lamda.nju.edu.cn}, \texttt{qyjiang24@gmail.com}, \texttt{liwujun@nju.edu.cn}}
\date{}
\maketitle
% \tableofcontents
% \newpage

\begin{abstract}
Due to its low storage cost and fast query speed, hashing has been widely used in \mbox{large-scale} image retrieval tasks. Hash bucket search returns data points within a given Hamming radius to each query, which can enable search at a constant or \mbox{sub-linear} time cost. However, existing hashing methods cannot achieve satisfactory retrieval performance for hash bucket search in complex scenarios, since they learn only one hash code for each image. More specifically, by using one hash code to represent one image, existing methods might fail to put similar image pairs to the buckets with a small Hamming distance to the query when the semantic information of images is complex. As a result, a large number of hash buckets need to be visited for retrieving similar images, based on the learned codes. This will deteriorate the efficiency of hash bucket search. In this paper, we propose a novel hashing framework, called \underline{m}ultiple \underline{c}ode \underline{h}ashing~(\modelname), to improve the performance of hash bucket search. The main idea of \modelname~is to learn multiple hash codes for each image, with each code representing a different region of the image. Furthermore, we propose a deep reinforcement learning algorithm to learn the parameters in MCH. To the best of our knowledge, this is the first work that proposes to learn multiple hash codes for each image in image retrieval. Experiments demonstrate that \modelname can achieve a significant improvement in hash bucket search, compared with existing methods that learn only one hash code for each image.
\end{abstract}
\section{Introduction}

%% intro ANN and hashing
With the rapid growth of multimedia applications~\cite{DBLP:journals/pami/TangSQLWYJ17,DBLP:conf/mm/FuJQSJCH18,DBLP:conf/cvpr/TomeiCBC19,DBLP:conf/iccv/QiZC019,DBLP:conf/mm/RagnarsdottirK19,DBLP:journals/pami/XuROWS19,DBLP:conf/kdd/YangZZX019,DBLP:conf/icdm/RazzakYYX19,DBLP:journals/tkde/YangFZLJ21,DBLP:journals/tkde/YangZWLXJ21,DBLP:conf/mm/0074ZGGZ22,DBLP:journals/titb/ZhangYYGLZYR22,DBLP:journals/toc/YangWZYXY22,DBLP:journals/tkdd/YangWSLZXY22,DBLP:journals/tkde/YangYBZZGXY23,DBLP:conf/aaai/YangHGXX23,DBLP:journals/chinaf/YangBGZYY23,DBLP:journals/tkde/YangYBZZGXY23,DBLP:journals/chinaf/YangBGZYY23,DBLP:journals/tgrs/MengWMYX24,DBLP:journals/tois/YangZSDZL24,DBLP:journals/datamine/LiYZ23}, large-scale and high-dimensional image data has brought much burden to the search engines. To ensure search efficiency, approximate nearest neighbor~(ANN)~\cite{DBLP:conf/vldb/GionisIM99,DBLP:conf/compgeom/DatarIIM04,DBLP:conf/focs/AndoniI06,DBLP:conf/ijcai/YangZXYZY21,DBLP:journals/fcsc/YangGLLLY24,DBLP:conf/icme/WanWGY24} search plays a fundamental role. As a popular solution of ANN search, hashing~\cite{DBLP:conf/vldb/GionisIM99,DBLP:conf/nips/KongL12,DBLP:journals/pami/GongLGP13,DBLP:conf/aaai/XiaPLLY14,DBLP:conf/cvpr/ShenSLS15,DBLP:conf/aaai/KangLZ16,DBLP:conf/ijcai/LiLDLG18,DBLP:conf/nips/SuZHT18,DBLP:journals/pami/WangZSSS18,DBLP:conf/mm/YeP18,DBLP:conf/mm/LiuLZWHJ18,DBLP:conf/mm/LiuNZCZY18,DBLP:conf/mm/MaoWZW18,DBLP:conf/cvpr/HeW019,DBLP:conf/mm/YanP0S0H19,DBLP:conf/mm/LiuNZY19,DBLP:conf/mm/HuWZP19,DBLP:conf/mm/LuZCLNZ19} has attracted much attention in recent years.

%% Hamming ranking and hash bucket search
The goal of hashing is to represent the data points as compact \mbox{similarity-preserving} binary hash codes in Hamming space~\cite{DBLP:conf/icml/NorouziF11,DBLP:conf/cvpr/ShenSSHT13,DBLP:conf/mm/ChenWLLNX19}. Based on hash code representations, the storage cost can be dramatically reduced. Furthermore, we can adopt two procedures, i.e., Hamming ranking and hash bucket search~\cite{DBLP:conf/icml/LiuWKC11,DBLP:conf/nips/LiuMKC14,DBLP:conf/icml/DaiGKHS17}, to accelerate search speed. The Hamming ranking procedure obtains a ranking list according to the Hamming distance between query and database points. The computational complexity of this procedure is $\OM(n)$~\cite{DBLP:journals/corr/Cai16b}, where $n$ is the number of database points. The hash bucket search procedure reorganizes the learned hash codes as a hash table~(index). Based on the hash index, hash bucket search can enable a \mbox{sub-linear} or even constant search by returning data points in those hash buckets whose Hamming distance to the query is smaller or equal to a given Hamming radius. In real applications, hash bucket search is typically more practical than Hamming ranking for fast search, especially for cases with a \mbox{large-scale} database.

\begin{figure}[tb]
\begin{center}
\includegraphics[scale=0.73]{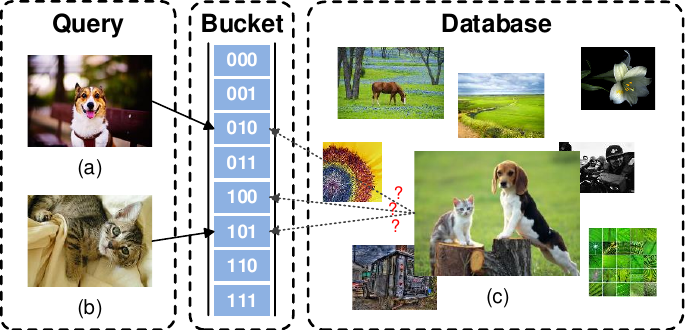}
\caption{An example for illustrating the shortcoming of existing hashing methods.}\vspace{-20pt}
\label{fig:introduction}
\end{center}
\end{figure}

%% problem of existing deep hashing methods
Over the past decades, many hashing methods~\cite{DBLP:journals/pami/GongLGP13,DBLP:conf/cvpr/ShenSLS15, DBLP:conf/cvpr/CaoLL018} have been proposed for image retrieval. However, all existing methods learn only one hash code for each image. When the semantic information of the images is complex, these methods may fail to put similar image pairs into the same hash bucket. An example for illustrating the shortcoming of existing hashing methods is shown in Figure~\ref{fig:introduction}. Suppose that there exists an image that contains both a dog and a cat in the database, i.e., image (c) in Figure~\ref{fig:introduction}. There are also two image queries that contain a dog and a cat separately, i.e., image (a) and image (b) in Figure~\ref{fig:introduction}. Suppose that we use $3$-bit binary hash codes to represent images. The category ``dog'' and category ``cat'' are represented by ``010'' and ``101'', respectively. For image (c), we can find that no matter which hash bucket it is mapped to, it cannot be directly retrieved~(without bit flipping) by these two queries at the same time. To get image~(c) as a search result for these two queries, we have to enlarge the search Hamming radius. For hash bucket search, the search cost will increase exponentially if we enlarge the search Hamming radius. In other words, the search efficiency will be significantly deteriorated. 

%% contributions
In this paper, we propose a novel hashing framework, called \underline{m}ultiple \underline{c}ode \underline{h}ashing~(\modelname), for image retrieval. The main contributions of this paper are summarized as follows:
\begin{itemize}
\item MCH learns multiple hash codes for each image, with each code representing a different region of the image. To the best of our knowledge, \modelname is the first hashing method that can learn multiple hash codes for each image. By representing each image with multiple hash codes, \modelname can keep the Hamming distance of similar image pairs small enough and enable efficient hash bucket search.
\item A novel deep reinforcement learning algorithm is proposed to learn the parameters in \modelname. In particular, an agent is trained in \modelname to explore different regions of the image that can better preserve the pairwise similarity than the whole image.
\item \modelname is a flexible framework that decomposes the multiple hash codes learning procedure into two steps: base hashing model learning step and agent learning step. This decomposition simplifies the learning procedure and enables the easy integration of different kinds of base hashing models.
\item Extensive experiments demonstrate that our \modelname can achieve a significant improvement in hash bucket search, compared with existing hashing methods that learn only one hash code for each image.
\end{itemize}

\section{Related Work}\label{sec:related-work}
In this section, we briefly review the related work, including hash bucket search and deep reinforcement learning.

\subsection{Hash Bucket Search}
Hash bucket search is an efficient method to locate the nearest codes in Hamming space. Given a hash code, it costs $O(1)$ time to locate its corresponding hash bucket. Ideally, if the data points in this bucket can satisfy the requirement of the search, the time cost of the hash bucket search is $O(1)$. If there are not enough data points in this bucket, we need to expand the Hamming radius by flipping bits to visit more hash buckets. In the worst case, all buckets are visited and the search cost will be $\OM(2^Q)$, where $Q$ denotes the hash code length. A general procedure of hash bucket search~\cite{DBLP:journals/corr/Cai16b} is summarized in Algorithm~\ref{alg:hash_bucket_search}. We can find that when the Hamming radius increases, the number of hash buckets need to be visited increases exponentially, which will severely deteriorate the retrieval efficiency in online search. 

\subsection{Deep Reinforcement Learning}
Reinforcement learning is a problem concerned with how an agent ought to learn behavior through \mbox{trial-and-error} interactions in a dynamic environment~\cite{DBLP:journals/jair/KaelblingLM96}. Due to the recent development of deep learning, deep reinforcement learning has attracted much attention and obtained progressive results~\cite{DBLP:journals/nature/MnihKSRVBGRFOPB15,DBLP:journals/nature/SilverHMGSDSAPL16}. More recently, deep reinforcement learning has been introduced to image hashing applications. Deep reinforcement learning approach for image hashing~(\mbox{DRLIH})~\cite{DBLP:journals/corr/abs-1802-02904} was proposed to learn hash functions in a sequential process. In~\cite{DBLP:conf/cvpr/DuanWLL018}, GraphBit was designed to mine the reliability of hash codes. These methods generate only one hash code for the whole image. None of them can generate multiple hash codes for modeling complex semantic information in images. Furthermore, these methods utilize deep reinforcement learning to generate \mbox{high-quality} hash codes, while our \modelname utilizes deep reinforcement learning to explore different regions of the image that can better preserve the pairwise similarity than the whole image.

\begin{algorithm}[t]
\caption{Hash bucket search algorithm}
\label{alg:hash_bucket_search}
\begin{algorithmic}
\REQUIRE
    Query $\q$, the hash function, hash index~(hash codes for all points in the database) and the number of required nearest neighbors $K$.
\ENSURE
    %$K$ points from the database which are close to $\q$.
    $K$ points closest to query $\q$ in the database.
\STATE {\bf Procedure}: Encode query $\q$ to hash code $\b$. Set Hamming radius $r=0$, result set $U=\varnothing$.
\WHILE{$|U|<K$}
    \STATE Get the bucket list $M$ with Hamming distance $r$ to $\b$.
    \FOR {each bucket $m$ in $M$}
        \STATE Put all the data points in $m$ to $U$.
        \IF{$|U| \geq K$}
            \STATE return the first $K$ data points in $U$.
        \ENDIF
    \ENDFOR
    \STATE $r = r + 1$
\ENDWHILE
\end{algorithmic}
\end{algorithm}

\section{Notation and Problem Definition}\label{sec:notation}
\subsection{Notation}
In this paper, we use boldface lowercase letters like $\w$ to denote vectors and boldface uppercase letters like $\W$ to denote matrices. $\Vert\w\Vert_2$ denotes the $L_2$-norm for vector $\w$. $W_{ij}$ denotes the element in the $i$-th row and $j$-th column of the matrix $\W$. $cos(\a, \b)$ denotes the the cosine distance between vector $\a$ and vector $\b$. $\sgn(\cdot)$ is an element-wise sign function which returns $1$ if the element is positive and returns $-1$ otherwise.
% defined as follows:

% $$ \sgn(x)=\left\{
% \begin{aligned}
% 1,\quad& x\geq 0 \\
% -1,\quad& x<0
% \end{aligned}
% \right.
% $$
% \begin{align}
% \sgn(x) = 
% \begin{cases}
% {1,} & {x\geq 0} \\ {-1,} & {x<0}
% \end{cases}
% \end{align}

\subsection{Problem Definition}
In this paper, we only focus on the setting with pairwise labels~\cite{DBLP:conf/cvpr/LiuWJJC12,DBLP:conf/iccv/CaoLWY17,DBLP:conf/cvpr/CaoLL018,DBLP:conf/iccv/Kang0L0Y19}. The technique in this paper can also be adapted to settings with other supervised information that includes pointwise labels~\cite{DBLP:conf/cvpr/ShenSLS15}, triplet labels~\cite{NINH:conf/cvpr/LaiPLY15,DRSH:journals/tip/ZhangLZZZ15} and ranking labels~\cite{RSH:conf/iccv/WangLSJ13,DBLP:conf/cvpr/0003CBS18}. This will be pursued in our future work. 

Suppose we have $N$ training samples which are denoted as $\X = \{\x_i\}_{i=1}^N$. Furthermore, the pairwise labels of the training set are also provided. The pairwise labels are denoted as: $\S=\{s_{ij}\}, s_{ij}\in\{0,1\}$, where $s_{ij}=1$ if $\x_i$ and $\x_j$ are similar, otherwise $s_{ij}=0$ if $\x_i$ and $\x_j$ are dissimilar.

The goal of \modelname is to learn a hash function \mbox{$g: \x_i\rightarrow\{\b_i^m\}_{m=1}^{t_i}$}, where $t_i$ denotes the number of hash codes\footnote{We represent the hash code as a vector form of $\{-1,+1\}^Q$ for convenience of learning. After learning, we can easily transform the learned hash code to the form of $\{0,1\}^Q$.} and each $\b_i^m \in \{-1, +1\}^Q$. The hash function $g$ encodes each data point $\x_i$ into $t_i$ $Q$-bit hash codes, which aims to represent the complex semantic information in the images to enable efficient hash bucket search.

\begin{figure*}[!htb]
\begin{center}
\includegraphics[scale=0.74]{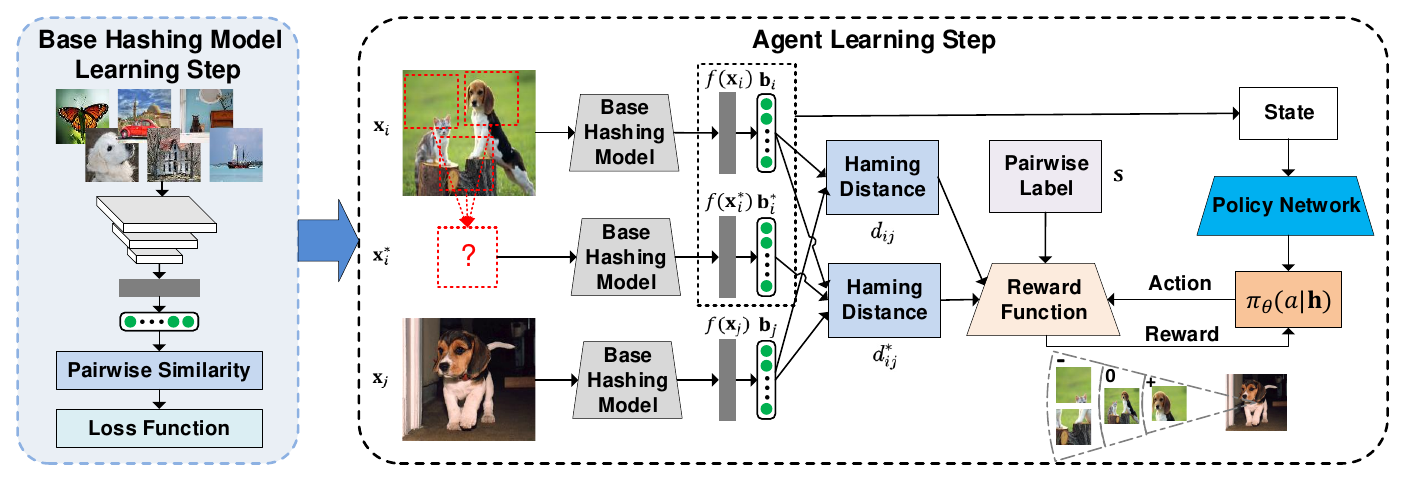}
\caption{An illustration of our \modelname framework, which consists of two steps. The first step is called base hashing model learning step, which learns a hash function through similarity-preserving learning. The second step is called agent learning step, which learns a decision strategy about multiple hash codes for image $\x_i$ through exploring different regions $\x_i^*$ that can better preserve the pairwise similarity than the whole image. Best viewed in color.}
\label{fig:framework}
\end{center}
% \vspace{-10pt}
\end{figure*}

\begin{table*}[tb]
\centering\scriptsize
\caption{The loss function and relaxation method for different hashing methods.}
% \vspace{-10pt}
\label{tab:hash_algorithm}
% \large{
\begin{tabular}{l|c|c|c}
\toprule
Method & Loss Function $L(\cdot, \cdot)$ & Relaxation Method & Regularization Term \\
\midrule
KSH~\cite{DBLP:conf/cvpr/LiuWJJC12}     & $(\frac{1}{Q}(Q - 2d_{ij}) - (2s_{ij} - 1))^2$  & Sigmoid Smoothing  & N/A \\
\hline
HashNet~\cite{DBLP:conf/iccv/CaoLWY17}      & $w_{ij}(\log(1 + \exp(\alpha(Q-2d_{ij})))-s_{ij}\alpha(Q-2d_{ij}))  $  & Tanh Smoothing & N/A  \\
\hline
ADSH~\cite{DBLP:conf/aaai/JiangL18}    & $((Q - 2d_{ij}) - Q(2s_{ij} - 1))^2$  & Continuous Relaxation & $\Vert \b_i - \u_i \Vert_2^2$ \\
\hline
DCH~\cite{DBLP:conf/cvpr/CaoLL018}      & $w_{ij}(s_{ij}\log(\frac{d_{ij}}{\gamma})+\log(1+\frac{\gamma}{d_{ij}}))$  & Continuous Relaxation & $\log(1+\frac{Q}{2\gamma}(1-\cos(\b_i,\u_i)))$  \\
\hline
MMHH~\cite{DBLP:conf/iccv/Kang0L0Y19}              & $w_{ij}(s_{ij}\log(1+\max(0, d_{ij} - H)) + (1-s_{ij})\log(1+\frac{1}{\max(H, d_{ij})}))$ &  Continuous Relaxation &  $\Vert \b_i - \u_i \Vert_2^2$ \\
\bottomrule
\end{tabular}%}
\end{table*}

\section{Multiple Code Hashing}
In this section, we present the details about our \modelname framework, including base hashing model learning step and agent learning step, which is illustrated in Figure~\ref{fig:framework}. 
% The purpose of the base hashing model learning step is to learn a hash function through \mbox{similarity-preserving} learning. The agent learning step aims to learn a decision strategy of multiple hash codes for an image by exploring different regions of the image that can better preserve the pairwise similarity than the whole image.

% \subsection{Model}\label{sec:mch}

\subsection{Base Hashing Model Learning Step}
The base hashing model in MCH is used to learn one hash code for a region of the image or the whole image. Both shallow hashing models and deep hashing models can be used as the base hashing model in MCH. Given an image $\x_i$, a shallow hashing model can first extract its hand-crafted features $f(\x_i)$ or utilize a backbone network to extract its features $f(\x_i)$ and then learn a linear or \mbox{non-linear} mapping followed by $\sgn(\cdot)$ function to get its hash code $\b_i$. A deep hashing model can use raw pixels in image $\x_i$ as the input and get its hash code $\b_i$ by \etoe representation learning and hash coding. The Hamming distance between hash codes $\b_i$ and $\b_j$ can be calculated as follows:
% {\setlength\abovedisplayskip{0.2cm}
% \setlength\abovedisplayskip{0.2cm}
\begin{align}
d_{ij} = \frac{1}{2}(Q - \b_i^\mathrm{T}\b_j).
\end{align}%}
To preserve the similarity between the data points, the Hamming distance between hash codes $\b_i$ and $\b_j$ should be relatively small if $s_{ij}=1$, while relatively large if $s_{ij}=0$. In other words, the goal of a hashing model is to solve the following problem:
\begin{align}
\min_{\Phi} \mathcal{L} = \sum_{i=1}^{N}\sum_{j=1}^{N} L(d_{ij}, s_{ij}),
\label{obj:pairwise_loss}
\end{align}
where $L(\cdot, \cdot)$ is a loss function and $\Phi$ denotes the parameters in $L(\cdot, \cdot)$ that need to be learned. Our \modelname is flexible enough to integrate different base hashing models with different types of loss functions $L(\cdot, \cdot)$. We adopt five different kinds of existing hashing methods as the base hashing models in this paper, which are listed in Table~\ref{tab:hash_algorithm}. Other existing hashing methods can also be adopted as base hashing models in MCH in a similar way as those in Table~\ref{tab:hash_algorithm}, which will not be further discussed because this is not the focus of this paper.

All the loss functions in Table~\ref{tab:hash_algorithm} are defined on the Hamming distance of data pairs. \mbox{Kernel-based} supervised hashing~(KSH)~\cite{DBLP:conf/cvpr/LiuWJJC12} uses $\ell_{2}$ loss function. Since the discrete optimization problem is hard to solve, KSH replaces $\sgn(\cdot)$ with Sigmoid function to get relaxed hash codes. HashNet~\cite{DBLP:conf/iccv/CaoLWY17}, asymmetric deep supervised hashing~(ADSH), deep Cauchy hashing~(DCH)~\cite{DBLP:conf/cvpr/CaoLL018} and \mbox{maximum-margin} Hamming hashing~(MMHH)~\cite{DBLP:conf/iccv/Kang0L0Y19} are recently proposed deep hashing methods. Similar to KSH, HashNet utilizes Tanh function to get relaxed hash codes. ADSH, DCH and MMHH solve the discrete optimization problem by relaxing $\b_i$ to a continuous vector $\u_i$ with a constraint that $\b_i = \sgn(\u_i)$. Such relaxation will cause a quantization error and reduce the quality of hash codes. To learn \mbox{high-quality} hash codes, the regularization term should be taken into consideration in the optimization problem. $\w_{ij}$ in the loss functions of HashNet, DCH and MMHH is the weight for each training pair. $\alpha, \gamma, H$ in the loss functions are \mbox{hyper-parameters} set by the corresponding authors.

\subsection{Agent Learning Step}
The agent learning step is used to learn multiple hash codes for each image with the base hashing model and reinforcement learning. 
%By training the base hashing model, we can learn one hash code for the whole image. However, learning only one hash code cannot put similar image pairs to the buckets within a small Hamming ball in complex scenarios. This will deteriorate the efficiency of hash bucket search. Hence, it is necessary to learn multiple hash codes, each for a different region of the image.

As shown in Figure~\ref{fig:framework}, a region $\x_i^*$ is randomly cropped from image $\x_i$. Then the features and hash codes for both $\x_i$ and $\x_i^*$ are provided through the base hashing model. The goal of agent learning is to determine whether the hash code $\b_i^*$ should be kept as one of the multiple hash codes for image $\x_i$.

{\emph{State Space}:} Given an image $\x_i$ and a cropped region $\x_i^*$, the state vector $\h_i$ is the concatenation of the feature vectors $f(\x_i)$, $f(\x_i^*)$, and the hash codes $\b_i, \b_i^*$ of $\x_i$ and region $\x_i^*$. This process can be formulated as follows:
\begin{align}
\h_i = \left[\left[f(\x_i); f(\x_i^*)\right]; \left[\b_i;\b_i^*\right]\right],
\end{align}
where $\left[\y; \z\right]$ denotes the vector concatenation operation on $\y$ and $\z$. More specifically, suppose each image is embedded into a $4096$-dimensional feature vector and the length of the hash code is $Q$, then the dimension of the state vector is $(8192+2Q)$.

{\emph{Action Space}:} Given the current state vector $\h_i$, the agent aims to select one action to keep or discard the hash code $\b_i^*$. The action $a_i$ has the following two possible choices: 
\begin{align}
a_i = 
\begin{cases}
{0,} & {\textrm{keep the hash code } \b_i} \\ {1,} & {\textrm{discard the hash code } \b_i}
\end{cases}
\end{align}

{\emph{Reward Function}:} Given the sampled action $a_i$, the reward function can be calculated based on the loss function $L(\cdot,\cdot)$ defined in~(\ref{obj:pairwise_loss}). First, we redefine the asymmetric Hamming distance from $\x_i$ to $\x_j$ when keeping the hash code $\b_i^*$ as follows:
\begin{align}
d^{*}_{ij} = \min\left(\frac{1}{2}(Q - \b_i^\mathrm{T}\b_j), \frac{1}{2}(Q - \b_i^{*\mathrm{T}}\b_j)\right),
\end{align}
which is consistent with the procedure of the hash bucket search. We can treat $\x_j$ as the query, $\x_i$ as a data point in the database and $\b_i, \b_i^*$ as two hash codes for $\x_i$ in Algorithm~\ref{alg:hash_bucket_search}. In other words, $\x_i$ is located in two different hash buckets, and the hash bucket with a smaller Hamming distance to $\b_j$ will be visited first, which indicates that $\x_i$ will be successfully retrieved. We define the following reward function to reflect the influence of the hash code $\b_i^*$ on the similarity preservation of data point $\x_i$.
\begin{align}
R(\h_i, a_i) = 
\begin{cases}
{0,} & {a_i = 0} \\ {\sum_{j=1}^{N}\left( L(d_{ij}, s_{ij}) - L(d^{*}_{ij}, s_{ij})\right),} & {a_i=1}
\end{cases}
\label{eq:reward}
\end{align}
If the agent discards the hash code $\b_i^*$~($a_i=0$), the reward is $0$ which means it does not affect the hash bucket search. When the agent keeps $\b_i^*$~($a_i=1$), the reward will be positive if the newly kept hash code $\b_i^*$ can better represent the semantic information of image $\x_i$ and lead to a lower loss for similarity preservation. Otherwise, the reward will be negative to force the agent to discard the meaningless hash code.
% In the environment, a region $\x^{*}_i$ is randomly cropped from image $\x_i$ as the training data for the agent at each step. The goal of the agent is to learn whether $\x^{*}_i$ should be preserved. For example, The ``grass'' region has a greater similarity to image $\x_j$ than the original image $\x_i$, which is contrary to their pairwise label $s_{ij}=0$. The agent will receive a negative reward to punish preserving this region. The ``cat'' or ``dog'' region has a smaller similarity to image $\x_j$, thus the agent will receive a positive reward if preserving these regions.

{\emph{Policy Network}:} We employ a multi-layer perceptron~(MLP) with a Softmax layer in the end as our policy network. More specifically, the policy network has five \mbox{fully-connected} layers. We apply ReLU as the activation function. We also perform Batch Normalization~\cite{DBLP:conf/icml/IoffeS15} to accelerate the training of the policy network. The input of the policy network is the state vector $\h_i$ and the output is the predicted distribution of the action $a_i$, which is denoted as $\pi_{\theta}(a_i | \h_i)$, where $\theta$ is the parameter of the policy network.

The goal of the policy network is to maximize the expected reward for the multiple hash code decision processes. We utilize the REINFORCE algorithm~\cite{DBLP:journals/ml/Williams92} to obtain the gradient of $\theta$, which is formulated as follows:
\begin{equation}
\begin{split}
\nabla_{\theta} \mathcal{Z}(\theta) & = \nabla_{\theta} \mathbb{E}_{\h, a} \left[ R(\h, a) \right] \\
& = \int_{\h, a} p(\h) \pi_{\theta}(a | \h) \nabla_{\theta} \log \pi_{\theta}(a | \h) R(\h, a) \\
& = \mathbb{E}_{\h, a} \left[\nabla_{\theta} \log \pi_{\theta}(a | \h) R(\h, a) \right] \\
& \approx \nabla_{\theta} \frac{1}{N} \sum_{i = 1}^{N} \log \pi_{\theta}(a_i | \h_i) R(\h_i, a_i).
\end{split}
\label{obj:policy}
\end{equation}

In \modelname, the parameters need to be learned contain $\Phi$ and $\theta$. We first learn the parameter $\Phi$ by training the base hashing model. Then in order to maximize the expected reward in~(\ref{obj:policy}), we adopt a \mbox{back-propagation} algorithm to learn the parameter $\theta$. The overall learning procedure is summarized in Algorithm~\ref{alg:mch}.

\begin{algorithm}[t]
\caption{Learning algorithm for \modelname}
\label{alg:mch}
\begin{algorithmic}
\REQUIRE% ~~\\
    $N$ Training images $\X=\{\x_i\}_{i=1}^N$.
\REQUIRE 
    Pairwise similarity labels $\S=\{s_{ij}\}_{i,j=1}^N$.
\ENSURE% ~~\\
    The parameter $\Phi$ and $\theta$.
\STATE {\bf Procedure}: Initialize parameter $\Phi$ and $\theta$, total iteration number $T$, mini-batch size $B$ and the base hashing model.
\STATE Learn the base hashing model according to its loss function and relaxation method in Table~\ref{tab:hash_algorithm}.
\FOR {$t=1\to T$}
    \STATE Randomly sample a mini-batch $\BM$ from $\{\x_i\}_{i=1}^N$.
    \STATE $\forall \x_i \in \BM$, calculate $f(\x_i)$ and $\b_i$ by forward propagation.
    \STATE $\forall \x_i \in \BM$, randomly crop a region $\x_i^*$, and calculate $f(\x_i^*)$ and $\b_i^*$ by forward propagation.
    \STATE Concatenate the feature vectors $f(\x_i), f(\x_i^*)$ and hash codes $\b_i, \b_i^*$ to generate state vector $\h_i$.
    \STATE For state vector $\h_i$, forward-propagate the policy network.
    \STATE Sample the action $a_i$ from the distribution $\pi_\theta(a_i|\h_i)$.
    \STATE Calculate the reward according to the reward function in~(\ref{eq:reward}).
    \STATE Update the parameter $\theta$ according to the gradient in~(\ref{obj:policy}).
\ENDFOR
\end{algorithmic}
\end{algorithm}

\subsection{Out-of-Sample Extension}\label{sec:out_of_sample}
After we have completed the whole learning procedure, we can only get the base hashing model and the agent. We still need to perform out-of-sample extension to predict the multiple hash codes for images unseen in the training set. 

For any new image $\x_i$ unseen in the training set, in addition to the original image, we crop $5$ regions~(four corners and one middle) from $\x_i$. The crop ratio is controlled by $\sigma \in [0, 1]$, which is a \mbox{hyper-parameter}. We use such a cropping strategy just for demonstrating the effectiveness of our \modelname in this paper. The cropping strategy can be adjusted according to actual needs. First, we utilize the learned base hashing model to obtain the hash codes of image $\x_i$ and all its cropped regions. Then we utilize the policy network to calculate the probability of keeping each hash code. Once the probability is larger than a given threshold of $\xi \in [0, 1]$, which is also a \mbox{hyper-parameter}, the corresponding hash code will be kept. Here, the probability of keeping the hash code for the original image $\x_i$ is fixed to $100\%$ to ensure that any image has at least one hash code.

\section{Experiments}\label{sec:exp}
In this section, we conduct extensive evaluations of the proposed method on three widely used benchmark datasets. To prove the effectiveness of \modelname, we adopt five different kinds of existing hashing methods as the base hashing models and illustrate the improvements brought by \modelname, in terms of both accuracy and efficiency.

\subsection{Datasets}
We select three widely used benchmark datasets for evaluation. They are \nuswide~\cite{DBLP:conf/civr/ChuaTHLLZ09}, \coco~\cite{DBLP:conf/eccv/LinMBHPRDZ14} and \flickr~\cite{DBLP:conf/mir/HuiskesTL10}.

\nuswide dataset contains $269,648$ images collected from the web, with each data point annotated with as least one class label from $81$ categories. Following~\cite{DBLP:conf/nips/LiSHT17}, we use the subset belonging to the $21$ most frequent classes. We randomly select $100$ images per class as the query set~($2,100$ images in total). The remaining images are used as the database set, from which we randomly sample $500$ images per class as the training set~($10,500$ images in total).
% \footnote{https://lms.comp.nus.edu.sg/research/NUS-WIDE.htm} 

\coco contains $82,783$ training images and $40,504$ validation images. Each image is annotated with some of the $80$ semantic concepts. Following~\cite{DBLP:conf/cvpr/CaoLL018}, we randomly select $5,000$ images as the query set. The remaining images are used as database set, and we randomly sample $10,000$ images from the database for training.
% \footnote{http://cocodataset.org/}

\flickr dataset contains $25,000$ images. Each image is annotated with multiple labels based on $21$ unique classes. 
% Following~\cite{DBLP:conf/ijcai/YangDLLT18}, 
We randomly select $5,000$ images as the query set. The remaining images are used as the database set, from which we randomly sample $10,000$ images as the training set.
% \footnote{https://press.liacs.nl/mirflickr/}

As the data point might belong to multiple categories, following~\cite{DBLP:conf/ijcai/LiWK16,DBLP:conf/nips/LiSHT17,DBLP:conf/cvpr/CaoLL018}, two data points that have at least one common semantic label are considered similar.

%---------------figure starts---------------
\begin{figure*}\centering
\begin{tikzpicture}[scale=0.85]
  \begin{groupplot}[legend cell align={left},
    group style={
      group size=3 by 1,
      x descriptions at=edge bottom,
    %   y descriptions at=edge left,
    },
    height=5cm,width=6.5cm,/tikz/font=\small,
    xlabel=Number of Bits,
    ]
    \nextgroupplot[title={NUS-WIDE},xtick={16,32,48,64},ytick={0,0.1,0.2,0.3,0.4,0.5,0.6},ymax=0.6,ylabel=Recall,ylabel style={yshift=-8pt}]
    \addplot [color=black!40!green,mark=o, mark size=1.5pt,mark options={fill=white}, line width=0.85pt]coordinates{(16,0.2615)(32,0.1867)(48,0.1523)(64,0.1500)};
    \addplot [color=black!40!green,dashed,mark=o, mark size=1.5pt,mark options={fill=white}, line width=0.85pt]coordinates{(16,0.1800)(32,0.1230)(48,0.1030)(64,0.1004)};
    \addplot [color=black!10!red,mark=square, mark size=1.5pt,mark options={fill=white}, line width=0.85pt]coordinates{(16,0.3309)(32,0.2514)(48,0.2284)(64,0.2003)};
    \addplot [color=black!10!red,dashed,mark=square, mark size=1.5pt,mark options={fill=white}, line width=0.85pt]coordinates{(16,0.2051)(32,0.1413)(48,0.1222)(64,0.1097)};
    \addplot [color=black!40!blue,mark=o, mark size=1.5pt,mark options={fill=white}, line width=0.85pt]coordinates{(16,0.3801)(32,0.2641)(48,0.2083)(64,0.1856)};
    \addplot [color=black!40!blue,dashed,mark=o, mark size=1.5pt,mark options={fill=white}, line width=0.85pt]coordinates{(16,0.2585)(32,0.1408)(48,0.1054)(64,0.0988)};
    \addplot [color=orange,mark=square, mark size=1.5pt,mark options={fill=white}, line width=0.85pt]coordinates{(16,0.4433)(32,0.4961)(48,0.5305)(64,0.5458)};
    \addplot [color=orange,dashed,mark=square, mark size=1.5pt,mark options={fill=white}, line width=0.85pt]coordinates{(16,0.4053)(32,0.4404)(48,0.4706)(64,0.4843)};
    \addplot [color=black,mark=o, mark size=1.5pt,mark options={fill=white}, line width=0.85pt]coordinates{(16,0.4660)(32,0.5308)(48,0.5333)(64,0.5464)};
    \addplot [color=black,dashed,mark=o, mark size=1.5pt,mark options={fill=white}, line width=0.85pt]coordinates{(16,0.4365)(32,0.4571)(48,0.4816)(64,0.4884)};
    \nextgroupplot[title={MS-COCO},xtick={16,32,48,64},ytick={0,0.1,0.2,0.3,0.4,0.5,0.6},ymax=0.6]
    \addplot [color=black!40!green,mark=o, mark size=1.5pt,mark options={fill=white}, line width=0.85pt]coordinates{(16,0.2237)(32,0.1446)(48,0.0906)(64,0.0808)};
    \addplot [color=black!40!green,dashed,mark=o, mark size=1.5pt,mark options={fill=white}, line width=0.85pt]coordinates{(16,0.0935)(32,0.0436)(48,0.0202)(64,0.01454)};
    \addplot [color=black!10!red,mark=square, mark size=1.5pt,mark options={fill=white}, line width=0.85pt]coordinates{(16,0.3810)(32,0.3469)(48,0.3344)(64,0.3344)};
    \addplot [color=black!10!red,dashed,mark=square, mark size=1.5pt,mark options={fill=white}, line width=0.85pt]coordinates{(16,0.2665)(32,0.2340)(48,0.2204)(64,0.2411)};
    \addplot [color=black!40!blue,mark=o, mark size=1.5pt,mark options={fill=white}, line width=0.85pt]coordinates{(16,0.3555)(32,0.2673)(48,0.2546)(64,0.2286)};
    \addplot [color=black!40!blue,dashed,mark=o, mark size=1.5pt,mark options={fill=white}, line width=0.85pt]coordinates{(16,0.2419)(32,0.1415)(48,0.1380)(64,0.1161)};
    \addplot [color=orange,mark=square, mark size=1.5pt,mark options={fill=white}, line width=0.85pt]coordinates{(16,0.4759)(32,0.5073)(48,0.5244)(64,0.5355)};
    \addplot [color=orange,dashed,mark=square, mark size=1.5pt,mark options={fill=white}, line width=0.85pt]coordinates{(16,0.4632)(32,0.4918)(48,0.5000)(64,0.5049)};
    \addplot [color=black,mark=o, mark size=1.5pt,mark options={fill=white}, line width=0.85pt]coordinates{(16,0.4759)(32,0.5001)(48,0.5106)(64,0.5150)};
    \addplot [color=black,dashed,mark=o, mark size=1.5pt,mark options={fill=white}, line width=0.85pt]coordinates{(16,0.4621)(32,0.4819)(48,0.4866)(64,0.4852)};
    \nextgroupplot[title={MIR FLICKR},xtick={16,32,48,64},legend style={font=\tiny,at={(1.05,1)},anchor=north west},ytick={0,0.1,0.2,0.3,0.4,0.5,0.6},ymax=0.5]
    \addplot [color=black!40!green,mark=o, mark size=1.5pt,mark options={fill=white}, line width=0.85pt]coordinates{(16,0.0587)(32,0.0176)(48,0.0085)(64,0.0047)};
    \addplot [color=black!40!green,dashed,mark=o, mark size=1.5pt,mark options={fill=white}, line width=0.85pt]coordinates{(16,0.0211)(32,0.0060)(48,0.0027)(64,0.0015)};
    \addplot [color=black!10!red,mark=square, mark size=1.5pt,mark options={fill=white}, line width=0.85pt]coordinates{(16,0.1010)(32,0.0594)(48,0.0386)(64,0.0307)};
    \addplot [color=black!10!red,dashed,mark=square, mark size=1.5pt,mark options={fill=white}, line width=0.85pt]coordinates{(16,0.0483)(32,0.0279)(48,0.013967)(64,0.0112)};
    \addplot [color=black!40!blue,mark=o, mark size=1.5pt,mark options={fill=white}, line width=0.85pt]coordinates{(16,0.1527)(32,0.0796)(48,0.0518)(64,0.0357)};
    \addplot [color=black!40!blue,dashed,mark=o, mark size=1.5pt,mark options={fill=white}, line width=0.85pt]coordinates{(16,0.0704)(32,0.0323)(48,0.0220)(64,0.0135)};
    \addplot [color=orange,mark=square, mark size=1.5pt,mark options={fill=white}, line width=0.85pt]coordinates{(16,0.1528)(32,0.2472)(48,0.2930)(64,0.3467)};
    \addplot [color=orange,dashed,mark=square, mark size=1.5pt,mark options={fill=white}, line width=0.85pt]coordinates{(16,0.0829)(32,0.0905)(48,0.1098)(64,0.1363)};
    \addplot [color=black,mark=o, mark size=1.5pt,mark options={fill=white}, line width=0.85pt]coordinates{(16,0.2435)(32,0.3969)(48,0.4287)(64,0.4589)};
    \addplot [color=black,dashed,mark=o, mark size=1.5pt,mark options={fill=white}, line width=0.85pt]coordinates{(16,0.1650)(32,0.2263)(48,0.2480)(64,0.2867)};
    \legend{{MCH-KSH},MCH,MCH-HashNet,HashNet,MCH-ADSH,ADSH,MCH-DCH,DCH,MCH-MMHH,MMHH};
  \end{groupplot}
\end{tikzpicture}
\caption{Recall within Hamming radius $0$ on the three benchmark datasets.}
\label{fig:hamming_0_recall}
\end{figure*}
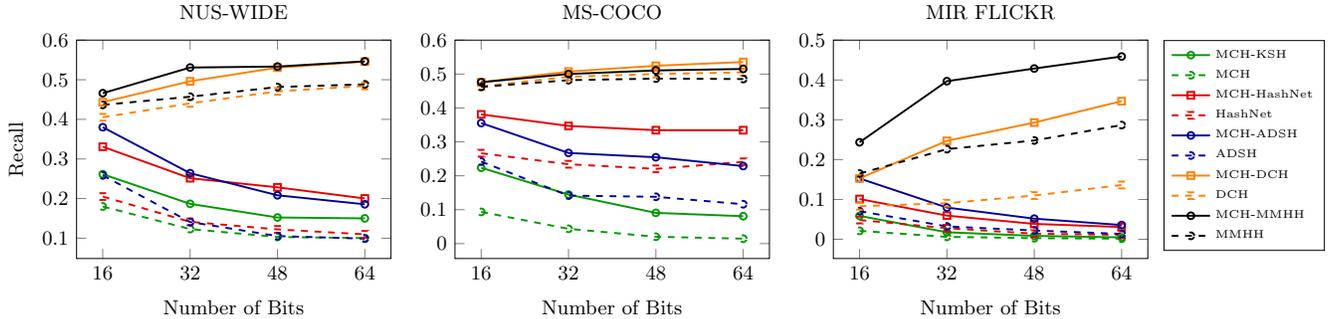
%---------------figure ends---------------

\subsection{Experimental Setup}
\subsubsection{Baselines and Evaluation Protocol}
Both shallow hashing methods and deep hashing methods are adopted as baselines for comparison. Shallow hashing methods include unsupervised and supervised methods. Unsupervised shallow hashing methods consist of locality-sensitive hashing~(LSH)~\cite{DBLP:conf/vldb/GionisIM99} and iterative quantization~(ITQ)~\cite{DBLP:journals/pami/GongLGP13}. Supervised shallow method is \mbox{kernel-based} supervised hashing~(KSH)~\cite{DBLP:conf/cvpr/LiuWJJC12}. Deep hashing methods consist of HashNet~\cite{DBLP:conf/iccv/CaoLWY17}, asymmetric deep supervised hashing~(ADSH), deep Cauchy hashing~(DCH)~\cite{DBLP:conf/cvpr/CaoLL018} and maximum-margin Hamming hashing~(MMHH)~\cite{DBLP:conf/iccv/Kang0L0Y19}. To demonstrate that our \modelname is an effective and flexible framework, we adopt both shallow hashing methods and deep hashing methods including KSH, HashNet, ADSH, DCH and MMHH as the base hashing model in \modelname. The corresponding MCH methods are called ``\modelname-KSH'',  ``\modelname-HashNet'', ``\modelname-ADSH'', ``\modelname-DCH'' and ``\modelname-MMHH'', respectively.

To verify that \modelname can enable the similar data points to fall into the Hamming ball within radius $0$, we adopt recall within Hamming radius $0$~(R@H$\mathbf{=0}$) and precision within Hamming radius $0$~(P@H$\mathbf{=0}$) to evaluate \modelname and baselines. We also adopt the widely used mean average precision~(mAP) to measure the accuracy of our \modelname and baselines. To evaluate the efficiency in the hash bucket search, we report \mbox{F1-bucket} curves to compare \modelname and baselines. More specifically, we conduct the hash bucket search with different settings of $K$, which is defined in Algorithm~\ref{alg:hash_bucket_search}. Then for each setting of $K$, we draw the F1 score and the number of hash buckets that need to be visited as a point. Finally, we connect these points in an ascending order of $K$ to get \mbox{F1-bucket} curves. Moreover, we report the average number of hash codes~(ANHC) for \modelname which reflects the average number of hash codes that MCH learns for an image. We can calculate the ANHC as follows:
\begin{align}
\mathrm{ANHC}(\{\x_i\}_{i=1}^N) = \frac{\sum_{i=1}^N\sum_{m=1}^{t_i}\mathbb{I}(P_{im}\geq\xi)}{N},
\end{align}
where $P_{im}$ denotes the probability that the $m$-th region of the $i$-th image $\x_i$ will be kept, $\xi$ is the threshold defined in Section~\ref{sec:out_of_sample}, and $\mathbb{I(\cdot)}$ is an indicator function. $\mathbb{I}(a) = 1$ if $a$ is true otherwise $0$.

\subsubsection{Implementation Details}
Our implementation of \modelname is based on PyTorch~\cite{DBLP:conf/nips/PaszkeGMLBCKLGA19}. For shallow hashing methods, we use the $4096$-dimensional features extracted by AlexNet~\cite{DBLP:conf/nips/KrizhevskySH12} \mbox{pre-trained} on ImageNet~\cite{DBLP:conf/cvpr/DengDSLL009} as image features. For deep hashing methods, we use original images as input and adopt AlexNet as the backbone architecture for a fair comparison. We \mbox{fine-tune} convolutional layers \emph{conv1-conv5} and \mbox{fully-connected} layers \emph{fc6-fc7} of AlexNet \mbox{pre-trained} on ImageNet and train the last hash layer, all through \mbox{back-propagation}. We use \mbox{mini-batch} stochastic gradient descent~(SGD) with $0.9$ momentum for training the policy network and choose the learning rate from $\left[10^{-5}, 10^{-3}\right]$. The initial learning rate is set to $1\times10^{-4}$ and the weight decay parameter is set to $5\times10^{-4}$. The total iteration number $T$ is set to $100$. The size of each \mbox{mini-batch} $B$ is set to $256$. The \mbox{hyper-parameters} of the base hashing model are set by following the suggestions of the corresponding authors. All the \mbox{hyper-parameters} shared by \modelname and its base hashing model are kept the same for a fair comparison. We select the unique \mbox{hyper-parameters} of our \modelname  $\sigma=0.5, \xi=0.5$ based on results of a validation strategy.

All experiments are run on a workstation with Intel(R) Xeon(R) Gold 6240 CPU@$2.60$GHz of $18$ cores, $384$G RAM and $8$ GeForce RTX 2080 Ti GPU cards. Furthermore, all experiments are run for five times with different random seeds and the average value is reported.

%---------------figure starts---------------
\begin{figure*}\centering
\begin{tikzpicture}[scale=0.85]
  \begin{groupplot}[legend cell align={left},
    group style={
      group size=3 by 1,
      x descriptions at=edge bottom,
    %   y descriptions at=edge left,
    },
    height=5cm,width=6.5cm,/tikz/font=\small,
    xlabel=Number of Bits,
    ]
    \nextgroupplot[title={NUS-WIDE},xtick={16,32,48,64},ytick={0,0.1,0.2,0.3,0.4,0.5,0.6,0.7,0.8},ymax=0.8,ylabel=Precision,ylabel style={yshift=-8pt}]
    \addplot [color=black!40!green,mark=o, mark size=1.5pt,mark options={fill=white}, line width=0.85pt]coordinates{(16,0.7573)(32,0.7299)(48,0.6811)(64,0.6483)};
    \addplot [color=black!40!green,dashed,mark=o, mark size=1.5pt,mark options={fill=white}, line width=0.85pt]coordinates{(16,0.7535)(32,0.7118)(48,0.6395)(64,0.5793)};
    \addplot [color=black!10!red,mark=square, mark size=1.5pt,mark options={fill=white}, line width=0.85pt]coordinates{(16,0.7664)(32,0.7094)(48,0.6709)(64,0.6376)};
    \addplot [color=black!10!red,dashed,mark=square, mark size=1.5pt,mark options={fill=white}, line width=0.85pt]coordinates{(16,0.7649)(32,0.6884)(48,0.6208)(64,0.5738)};
    \addplot [color=black!40!blue,mark=o, mark size=1.5pt,mark options={fill=white}, line width=0.85pt]coordinates{(16,0.7751)(32,0.7684)(48,0.7394)(64,0.7080)};
    \addplot [color=black!40!blue,dashed,mark=o, mark size=1.5pt,mark options={fill=white}, line width=0.85pt]coordinates{(16,0.7755)(32,0.7642)(48,0.7216)(64,0.6817)};
    \addplot [color=orange,mark=square, mark size=1.5pt,mark options={fill=white}, line width=0.85pt]coordinates{(16,0.7452)(32,0.7078)(48,0.6893)(64,0.6800)};
    \addplot [color=orange,dashed,mark=square, mark size=1.5pt,mark options={fill=white}, line width=0.85pt]coordinates{(16,0.7469)(32,0.7020)(48,0.6780)(64,0.6697)};
    \addplot [color=black,mark=o, mark size=1.5pt,mark options={fill=white}, line width=0.85pt]coordinates{(16,0.7431)(32,0.7183)(48,0.7036)(64,0.7010)};
    \addplot [color=black,dashed,mark=o, mark size=1.5pt,mark options={fill=white}, line width=0.85pt]coordinates{(16,0.7441)(32,0.7111)(48,0.6927)(64,0.6894)};
    \nextgroupplot[title={MS-COCO},xtick={16,32,48,64},ytick={0,0.1,0.2,0.3,0.4,0.5,0.6,0.7},ymax=0.7]
    \addplot [color=black!40!green,mark=o, mark size=1.5pt,mark options={fill=white}, line width=0.85pt]coordinates{(16,0.6476)(32,0.6594)(48,0.6166)(64,0.5805)};
    \addplot [color=black!40!green,dashed,mark=o, mark size=1.5pt,mark options={fill=white}, line width=0.85pt]coordinates{(16,0.6375)(32,0.6108)(48,0.5354)(64,0.4842)};
    \addplot [color=black!10!red,mark=square, mark size=1.5pt,mark options={fill=white}, line width=0.85pt]coordinates{(16,0.6656)(32,0.6086)(48,0.5588)(64,0.5335)};
    \addplot [color=black!10!red,dashed,mark=square, mark size=1.5pt,mark options={fill=white}, line width=0.85pt]coordinates{(16,0.6617)(32,0.5994)(48,0.5475)(64,0.5207)};
    \addplot [color=black!40!blue,mark=o, mark size=1.5pt,mark options={fill=white}, line width=0.85pt]coordinates{(16,0.6274)(32,0.6106)(48,0.5604)(64,0.5250)};
    \addplot [color=black!40!blue,dashed,mark=o, mark size=1.5pt,mark options={fill=white}, line width=0.85pt]coordinates{(16,0.6152)(32,0.5891)(48,0.5296)(64,0.4866)};
    \addplot [color=orange,mark=square, mark size=1.5pt,mark options={fill=white}, line width=0.85pt]coordinates{(16,0.6539)(32,0.6315)(48,0.6215)(64,0.6183)};
    \addplot [color=orange,dashed,mark=square, mark size=1.5pt,mark options={fill=white}, line width=0.85pt]coordinates{(16,0.6491)(32,0.6258)(48,0.6164)(64,0.6116)};
    \addplot [color=black,mark=o, mark size=1.5pt,mark options={fill=white}, line width=0.85pt]coordinates{(16,0.6418)(32,0.6262)(48,0.6123)(64,0.6041)};
    \addplot [color=black,dashed,mark=o, mark size=1.5pt,mark options={fill=white}, line width=0.85pt]coordinates{(16,0.6353)(32,0.6179)(48,0.6068)(64,0.6003)};
    \nextgroupplot[title={MIR FLICKR},xtick={16,32,48,64},legend style={font=\tiny,at={(1.05,1)},anchor=north west},ytick={0,0.1,0.2,0.3,0.4,0.5,0.6,0.7,0.8,0.9},ymax=0.9]
    \addplot [color=black!40!green,mark=o, mark size=1.5pt,mark options={fill=white}, line width=0.85pt]coordinates{(16,0.8562)(32,0.7456)(48,0.6554)(64,0.5776)};
    \addplot [color=black!40!green,dashed,mark=o, mark size=1.5pt,mark options={fill=white}, line width=0.85pt]coordinates{(16,0.8293)(32,0.6512)(48,0.4908)(64,0.3684)};
    \addplot [color=black!10!red,mark=square, mark size=1.5pt,mark options={fill=white}, line width=0.85pt]coordinates{(16,0.8650)(32,0.7449)(48,0.6554)(64,0.5844)};
    \addplot [color=black!10!red,dashed,mark=square, mark size=1.5pt,mark options={fill=white}, line width=0.85pt]coordinates{(16,0.8256)(32,0.6301)(48,0.5036)(64,0.4118)};
    \addplot [color=black!40!blue,mark=o, mark size=1.5pt,mark options={fill=white}, line width=0.85pt]coordinates{(16,0.8731)(32,0.8186)(48,0.7783)(64,0.7389)};
    \addplot [color=black!40!blue,dashed,mark=o, mark size=1.5pt,mark options={fill=white}, line width=0.85pt]coordinates{(16,0.8669)(32,0.7961)(48,0.7317)(64,0.6655)};
    \addplot [color=orange,mark=square, mark size=1.5pt,mark options={fill=white}, line width=0.85pt]coordinates{(16,0.8800)(32,0.8567)(48,0.8433)(64,0.8339)};
    \addplot [color=orange,dashed,mark=square, mark size=1.5pt,mark options={fill=white}, line width=0.85pt]coordinates{(16,0.8679)(32,0.8246)(48,0.8099)(64,0.7977)};
    \addplot [color=black,mark=o, mark size=1.5pt,mark options={fill=white}, line width=0.85pt]coordinates{(16,0.8758)(32,0.8498)(48,0.8407)(64,0.8319)};
    \addplot [color=black,dashed,mark=o, mark size=1.5pt,mark options={fill=white}, line width=0.85pt]coordinates{(16,0.8653)(32,0.8342)(48,0.8212)(64,0.8173)};
    \legend{{MCH-KSH},MCH,MCH-HashNet,HashNet,MCH-ADSH,ADSH,MCH-DCH,DCH,MCH-MMHH,MMHH};
  \end{groupplot}
\end{tikzpicture}
\caption{Precision within Hamming radius $0$ on the three benchmark datasets.}
\label{fig:hamming_0_precision}
\end{figure*}
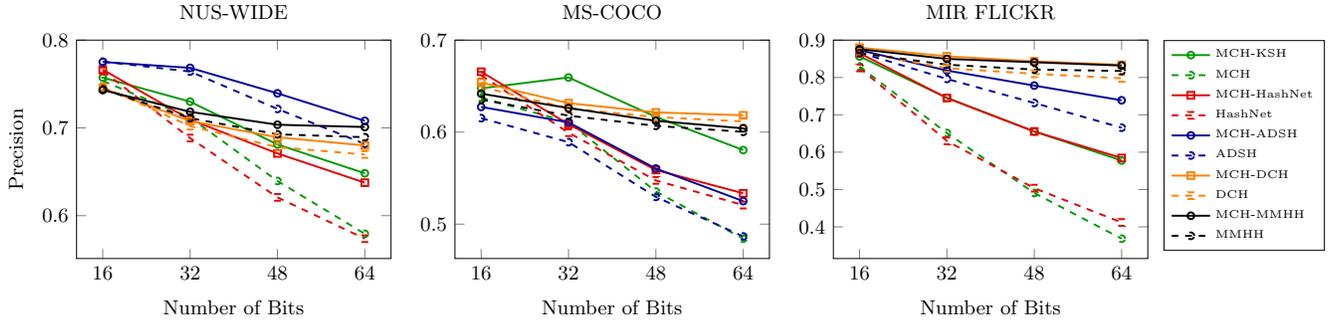
%---------------figure ends---------------

\begin{table*}[tb]
\centering\scriptsize
\caption{mAP on three benchmark datasets with different code lengths. Results that are improved by our \modelname compared to the base hashing model are shown in boldface.}
\label{tab:map}
\begin{tabular}{lcccccccccccc}
\toprule
\multirow{2}{*}{Method}    & \multicolumn{4}{c}{\nuswide} & \multicolumn{4}{c}{\coco} & \multicolumn{4}{c}{\flickr} \\
\cmidrule(lr){2-5} \cmidrule(lr){6-9} \cmidrule(lr){10-13}
    & 16 bits & 32 bits & 48 bits & 64 bits & 16 bits & 32 bits & 48 bits & 64 bits & 16 bits & 32 bits & 48 bits & 64 bits  \\
\midrule
LSH & {0.3449} & {0.3653} & {0.3815} & {0.3861} & {0.3660} & {0.3763} & {0.3880} & {0.3947} & {0.5710} & {0.5799} & {0.5908} & {0.5931} \\
ITQ & {0.5183} & {0.5140} & {0.5172} & {0.5185} & {0.4589} & {0.4711} & {0.4788} & {0.4826} & {0.6479} & {0.6516} & {0.6536} & {0.6544} \\
% IsoH & {0.4915} & {0.4888} & {0.4901} & {0.4909} & {0.4427} & {0.4518} & {0.4586} & {0.4631} & {0.6400} & {0.6430} & {0.6432} & {0.6454} \\
\midrule
KSH & {0.6894} & {0.7003} & {0.7047} & {0.7062} & {0.5614} & {0.5768} & {0.5821} & {0.5866} & {0.7923} & {0.8005} & {0.8024} & {0.8038} \\
MCH-KSH & {\bf0.6955} & {\bf0.7062} & {\bf0.7104} & {\bf0.7128} & {\bf0.5691} & {\bf0.5858} & {\bf0.5915} & {\bf0.5967} & {\bf0.8051} & {\bf0.8110} & {\bf0.8117} & {\bf0.8114} \\
\midrule
% DPSH & {0.6903} & {0.6976} & {0.6963} & {0.6998} & {0.5734} & {0.5931} & {0.5959} & {0.6000} & {0.8543} & {0.8636} & {0.8634} & {0.8596} \\
% MCH-DPSH & {\bf0.6940} & {\bf0.7012} & {\bf0.6995} & {\bf0.7028} & {\bf0.5753} & {\bf0.5950} & {\bf0.5977} & {\bf0.6020} & {\bf0.8593} & {\bf0.8671} & {\bf0.8667} & {\bf0.8623} \\
HashNet & {0.7064} & {0.7184} & {0.7211} & {0.7225} & {0.5766} & {0.5859} & {0.5886} & {0.5846} & {0.8454} & {0.8515} & {0.8531} & {0.8540} \\
MCH-HashNet & {\bf0.7098} & {\bf0.7221} & {\bf0.7246} & {\bf0.7266} & {\bf0.5800} & {\bf0.5901} & {\bf0.5929} & {\bf0.5887} & {\bf0.8491} & {\bf0.8553} & {\bf0.8563} & {\bf0.8573} \\
ADSH & {0.7161} & {0.7391} & {0.7444} & {0.7461} & {0.5643} & {0.5934} & {0.6055} & {0.6135} & {0.8352} & {0.8452} & {0.8471} & {0.8470} \\
MCH-ADSH & {\bf0.7179} & {\bf0.7403} & {\bf0.7454} & {\bf0.7471} & {\bf0.5684} & {\bf0.5985} & {\bf0.6098} & {\bf0.6174} & {\bf0.8383} & {\bf0.8475} & {\bf0.8502} & {\bf0.8511} \\
DCH & {0.6804} & {0.6796} & {0.6765} & {0.6761} & {0.5542} & {0.5554} & {0.5550} & {0.5546} & {0.8273} & {0.8267} & {0.8251} & {0.8214} \\
MCH-DCH & {\bf0.6832} & {\bf0.6840} & {\bf0.6819} & {\bf0.6801} & {\bf0.5607} & {\bf0.5665} & {\bf0.5650} & {\bf0.5647} & {\bf0.8359} & {\bf0.8348} & {\bf0.8280} & {\bf0.8230} \\
MMHH & {0.6762} & {0.6762} & {0.6716} & {0.6702} & {0.5499} & {0.5498} & {0.5456} & {0.5442} & {0.8199} & {0.8120} & {0.8119} & {0.8091} \\
MCH-MMHH & {\bf0.6791} & {\bf0.6818} & {\bf0.6795} & {\bf0.6763} & {\bf0.5563} & {\bf0.5597} & {\bf0.5547} & {\bf0.5492} & {\bf0.8276} & {\bf0.8200} & {\bf0.8162} & {\bf0.8138} \\
\bottomrule
\end{tabular}
\end{table*} 

\input{figure/F1-16bits.tex}

\subsection{Accuracy}
% The performance of recall within Hamming radius $0$ is crucial for retrieving top-ranking data points by the hash bucket search, since all data points might be pruned out due to the highly sparse Hamming space. As shown in Figure~\ref{fig:hamming_0_recall}, \modelname achieves a significant improvement than its base hashing model on all benchmark datasets with regard to different code lengths. This validates that \modelname can make more data points fall in the Hamming ball within radius $0$ to the query and enable more efficient retrieval of top-ranking data points. The performance of precision within Hamming radius $0$ is also important for retrieving the top-ranking data points through the hash bucket search, since it only needs $O(1)$ time cost and enables efficient pruning. As shown in Figure~\ref{fig:hamming_0_precision}, \modelname achieves higher results than its base hashing model on all three benchmark datasets with regard to different code lengths. This validates that \modelname can enable more efficient retrieval of top-ranking data points without reducing accuracy. 

The recall and precision within Hamming radius $0$ reflect the performance for retrieving top-ranking data points by the hash bucket search at $O(1)$ time cost. The recall within Hamming radius $0$ is shown in Figure~\ref{fig:hamming_0_recall}. We can find that \modelname achieves much better results than its base hashing model on all benchmark datasets with regard to different code lengths. This verifies that \modelname can make more similar data points fall into the Hamming ball within radius $0$ to the query and enable more efficient retrieval of top-ranking data points. Figure~\ref{fig:hamming_0_precision} shows the precision within Hamming radius $0$. We can find that \modelname can also improve the precision within Hamming radius $0$, compared with its base hashing model. This verifies that \modelname will not reduce the accuracy of retrieving top-ranking data points.

\input{figure/F1-32bits.tex}

\input{figure/F1-64bits.tex}

The mAP results of all methods are listed in Table~\ref{tab:map}, which show that \modelname can achieve better accuracy than its base hashing model. Please note that our \modelname is mainly designed to improve the efficiency of hash bucket search. The improvement in mAP is to illustrate that our \modelname will not reduce accuracy. 

To verify the efficiency of MCH in hash bucket search, \mbox{F1-bucket} curves of all methods with different code lengths are illustrated in Figure~\ref{fig:f1_bucket_16}, Figure~\ref{fig:f1_bucket_32} and Figure~\ref{fig:f1_bucket_64}, respectively. The results show that \modelname can perform much more efficient hash bucket search than the base hashing model. Specifically, compared to KSH, the best shallow hashing method with deep features as input, \mbox{\modelname-KSH} can obtain the same F1 score while greatly reducing the number of hash buckets that need to be visited. The \mbox{F1-bucket} curves also show some interesting phenomenons.
\begin{inparaenum}[(1)]
\item Without modeling the pairwise similarity information, unsupervised shallow hashing methods are worse than supervised hashing methods.
\item ADSH treats query points and database points in an asymmetric way. In most situations, ADSH achieves more efficient hash bucket search, compared with symmetric supervised hashing methods.
\item Different symmetric supervised hashing methods have large distinctions in the efficiency of hash bucket search. The reason is that different types of loss functions impose distinct penalties for similar point pairs~($s_{ij}=1$)~\cite{DBLP:conf/cvpr/CaoLL018}. These observations may give some insights about designing new hashing methods for more efficient hash bucket search.
\end{inparaenum}
% can greatly reduce the number of hash buckets that need to be visited to achieve the same accuracy. In other words, \mbox{\modelname-KSH} achieves a higher F1 score when visiting the same number of hash buckets, compared with KSH. 

\subsection{Visualization Study}
\begin{figure*}[!htb]
\centering
\includegraphics[scale=0.475]{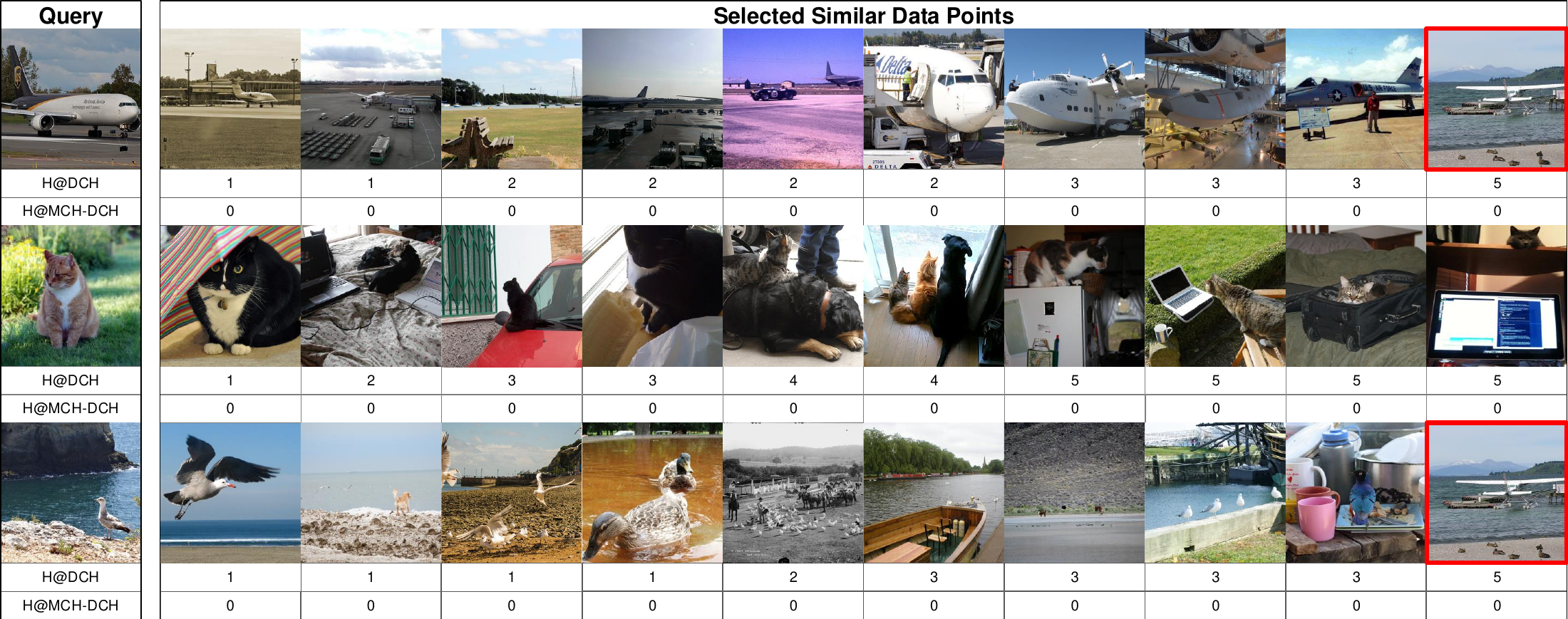}
\caption{Visualization study on \coco dataset. The first column denotes the query and the following $10$ columns denote the retrieved images. ``H@DCH'' denotes the Hamming distance between the query and retrieved images, which is obtained by DCH. ``H@MCH-DCH'' is defined similarly. The red box indicates that the same image is retrieved by queries of different categories.}
\label{fig:case_study}
\end{figure*}

We perform a visualization study on \coco dataset to give an intuition about why MCH can outperform existing methods. We randomly select some query images from the query set and select $10$ similar data points for each query from the database set. We choose DCH as the base hashing model of \modelname and set the code length to $32$ bits. Results are shown in Figure~\ref{fig:case_study}, in which the first column denotes the query, and the following $10$ columns denote the retrieved data points to the query. The Hamming distance between the query and each retrieved data point is also shown in Figure~\ref{fig:case_study}. The results show that our \modelname can better represent the data points with complex semantic information and enable more similar data points to fall into the Hamming ball within radius $0$ to the query. There also exists an image in the database set describing both the airplane and the bird. Our \modelname can enable it to appear in multiple hash buckets simultaneously. Hence, it can be retrieved in $O(1)$ time by queries of different categories. This validates that \modelname can enable more efficient hash bucket search.

\input{figure/hyper-param.tex}

\subsection{Sensitivity to Hyper-Parameter}
Since the hash bucket search efficiency is the main focus of this paper, we study the sensitivity of the \mbox{hyper-parameters} $\sigma$ and $\xi$ on the hash bucket search efficiency. The efficiency~(\mbox{F1-bucket}) of \mbox{MCH-DCH} and DCH with code length being $32$ bits are shown in Figure~\ref{fig:hyper_parameter}. From Figure~\ref{fig:hyper_parameter}~(a), we can see that \modelname is not sensitive to $\sigma$ in the range $0.3 \leq \sigma \leq 0.5$. To further analyze the sensitivity to $\sigma$, we select $131$ queries that include the ``Airplane'' category and report the \mbox{F1-bucket} curves in Figure~\ref{fig:hyper_parameter}~(b) using only these selected queries. We can see that \modelname is not sensitive to $\sigma$ in the range $0.4 \leq \sigma \leq 0.6$. We also select $141$ queries that include ``Bird'' category and results are shown in Figure~\ref{fig:hyper_parameter}~(c). We can find that the hyper-parameter sensitivity results are similar to those using all queries. From Figure~\ref{fig:hyper_parameter}~(d), we can see that \modelname is not sensitive to $\xi$ in the range $0.1 \leq \sigma \leq 0.9$. 

We also report the average number of hash codes of \mbox{MCH-DCH} in Table~\ref{tab:code_num}. Please note that when $\sigma = 1.0$, \mbox{MCH-DCH} will degenerate to its base hashing model DCH. We can find that a more efficient hash bucket search in Figure~\ref{fig:hyper_parameter} can be achieved with a larger average number of hash codes.

\begin{table}[!htb]
\centering
\caption{The effect of hyper-parameters $\sigma$ and $\xi$ on the average number of hash codes on \coco dataset.}
\label{tab:code_num}
% \large{
\begin{tabular}{c|cccccc}
\toprule
\diagbox[height=1.35\line]{$\sigma$}{$\xi$} & 0.1 & 0.3 & 0.5 & 0.7 & 0.9 & 1.0 \\
\midrule
0.1 & {1.12} & {1.08} & {1.06} & {1.05} & {1.03} & {1.00}  \\
0.3 & {1.79} & {1.68} & {1.62} & {1.56} & {1.49} & {1.04}  \\
0.5 & {1.68} & {1.59} & {1.54} & {1.50} & {1.43} & {1.05}  \\
0.7 & {1.47} & {1.40} & {1.36} & {1.33} & {1.28} & {1.01}  \\
0.9 & {1.43} & {1.37} & {1.34} & {1.31} & {1.26} & {1.02}  \\
1.0 & {1.00} & {1.00} & {1.00} & {1.00} & {1.00} & {1.00} \\
\bottomrule
\end{tabular}%}
\end{table}

\section{Conclusion}\label{sec:conclusion}
In this paper, we propose a novel hashing method, called multiple codes hashing~(\modelname), for efficient image retrieval. \modelname is the first hashing method that proposes to learn multiple hash codes for each image, with each code representing a different region of the image. Furthermore, we propose a deep reinforcement learning algorithm to learn the parameters in \modelname. \modelname provides a flexible framework that can enable the easy integration of different kinds of base hashing models. Extensive experiments demonstrate that our \modelname can achieve a significant improvement in hash bucket search, compared with existing hashing methods that learn only one hash code for each image.

\bibliographystyle{plain}
\bibliography{sample-base}

\begin{thebibliography}{10}

\bibitem{DBLP:conf/focs/AndoniI06}
Alexandr Andoni and Piotr Indyk.
\newblock Near-optimal hashing algorithms for approximate nearest neighbor in
  high dimensions.
\newblock In {\em {FOCS}}, pages 459--468, 2006.

\bibitem{DBLP:journals/corr/Cai16b}
Deng Cai.
\newblock A revisit of hashing algorithms for approximate nearest neighbor
  search.
\newblock {\em CoRR}, abs/1612.07545, 2016.

\bibitem{DBLP:conf/cvpr/CaoLL018}
Yue Cao, Mingsheng Long, Bin Liu, and Jianmin Wang.
\newblock Deep cauchy hashing for hamming space retrieval.
\newblock In {\em {CVPR}}, pages 1229--1237, 2018.

\bibitem{DBLP:conf/iccv/CaoLWY17}
Zhangjie Cao, Mingsheng Long, Jianmin Wang, and Philip~S. Yu.
\newblock Hashnet: Deep learning to hash by continuation.
\newblock In {\em {ICCV}}, pages 5609--5618, 2017.

\bibitem{DBLP:conf/mm/ChenWLLNX19}
Zhen{-}Duo Chen, Yongxin Wang, Hui{-}Qiong Li, Xin Luo, Liqiang Nie, and
  Xin{-}Shun Xu.
\newblock A two-step cross-modal hashing by exploiting label correlations and
  preserving similarity in both steps.
\newblock In {\em {MM}}, pages 1694--1702, 2019.

\bibitem{DBLP:conf/civr/ChuaTHLLZ09}
Tat{-}Seng Chua, Jinhui Tang, Richang Hong, Haojie Li, Zhiping Luo, and Yantao
  Zheng.
\newblock {NUS-WIDE:} a real-world web image database from national university
  of singapore.
\newblock In {\em {CIVR}}, pages 1--9, 2009.

\bibitem{DBLP:conf/icml/DaiGKHS17}
Bo~Dai, Ruiqi Guo, Sanjiv Kumar, Niao He, and Le~Song.
\newblock Stochastic generative hashing.
\newblock In {\em {ICML}}, pages 913--922, 2017.

\bibitem{DBLP:conf/compgeom/DatarIIM04}
Mayur Datar, Nicole Immorlica, Piotr Indyk, and Vahab~S. Mirrokni.
\newblock Locality-sensitive hashing scheme based on p-stable distributions.
\newblock In {\em {SCG}}, pages 253--262, 2004.

\bibitem{DBLP:conf/cvpr/DengDSLL009}
Jia Deng, Wei Dong, Richard Socher, Li{-}Jia Li, Kai Li, and Fei{-}Fei Li.
\newblock Imagenet: {A} large-scale hierarchical image database.
\newblock In {\em {CVPR}}, pages 248--255, 2009.

\bibitem{DBLP:conf/cvpr/DuanWLL018}
Yueqi Duan, Ziwei Wang, Jiwen Lu, Xudong Lin, and Jie Zhou.
\newblock Graphbit: Bitwise interaction mining via deep reinforcement learning.
\newblock In {\em {CVPR}}, pages 8270--8279, 2018.

\bibitem{DBLP:conf/mm/FuJQSJCH18}
Zhihang Fu, Zhongming Jin, Guo{-}Jun Qi, Chen Shen, Rongxin Jiang, Yaowu Chen,
  and Xian{-}Sheng Hua.
\newblock Previewer for multi-scale object detector.
\newblock In {\em {MM}}, pages 265--273, 2018.

\bibitem{DBLP:conf/vldb/GionisIM99}
Aristides Gionis, Piotr Indyk, and Rajeev Motwani.
\newblock Similarity search in high dimensions via hashing.
\newblock In {\em {VLDB}}, pages 518--529, 1999.

\bibitem{DBLP:journals/pami/GongLGP13}
Yunchao Gong, Svetlana Lazebnik, Albert Gordo, and Florent Perronnin.
\newblock Iterative quantization: {A} procrustean approach to learning binary
  codes for large-scale image retrieval.
\newblock {\em {TPAMI}}, 35(12):2916--2929, 2013.

\bibitem{DBLP:conf/cvpr/0003CBS18}
Kun He, Fatih {\c{C}}akir, Sarah~Adel Bargal, and Stan Sclaroff.
\newblock Hashing as tie-aware learning to rank.
\newblock In {\em {CVPR}}, pages 4023--4032, 2018.

\bibitem{DBLP:conf/cvpr/HeW019}
Xiangyu He, Peisong Wang, and Jian Cheng.
\newblock K-nearest neighbors hashing.
\newblock In {\em {CVPR}}, pages 2839--2848, 2019.

\bibitem{DBLP:conf/mm/HuWZP19}
Peng Hu, Xu~Wang, Liangli Zhen, and Dezhong Peng.
\newblock Separated variational hashing networks for cross-modal retrieval.
\newblock In {\em {MM}}, pages 1721--1729, 2019.

\bibitem{DBLP:conf/mir/HuiskesTL10}
Mark~J. Huiskes, Bart Thomee, and Michael~S. Lew.
\newblock New trends and ideas in visual concept detection: The {MIR} flickr
  retrieval evaluation initiative.
\newblock In {\em {MIR}}, pages 527--536, 2010.

\bibitem{DBLP:conf/icml/IoffeS15}
Sergey Ioffe and Christian Szegedy.
\newblock Batch normalization: Accelerating deep network training by reducing
  internal covariate shift.
\newblock In {\em {ICML}}, pages 448--456, 2015.

\bibitem{DBLP:conf/aaai/JiangL18}
Qing{-}Yuan Jiang and Wu{-}Jun Li.
\newblock Asymmetric deep supervised hashing.
\newblock In {\em {AAAI}}, pages 3342--3349, 2018.

\bibitem{DBLP:journals/jair/KaelblingLM96}
Leslie~Pack Kaelbling, Michael~L. Littman, and Andrew~W. Moore.
\newblock Reinforcement learning: {A} survey.
\newblock {\em JAIR}, 4:237--285, 1996.

\bibitem{DBLP:conf/iccv/Kang0L0Y19}
Rong Kang, Yue Cao, Mingsheng Long, Jianmin Wang, and Philip~S. Yu.
\newblock Maximum-margin hamming hashing.
\newblock In {\em {ICCV}}, pages 8251--8260, 2019.

\bibitem{DBLP:conf/aaai/KangLZ16}
Wang{-}Cheng Kang, Wu{-}Jun Li, and Zhi{-}Hua Zhou.
\newblock Column sampling based discrete supervised hashing.
\newblock In {\em {AAAI}}, pages 1230--1236, 2016.

\bibitem{DBLP:conf/nips/KongL12}
Weihao Kong and Wu{-}Jun Li.
\newblock Isotropic hashing.
\newblock In {\em NeurIPS}, pages 1655--1663, 2012.

\bibitem{DBLP:conf/nips/KrizhevskySH12}
Alex Krizhevsky, Ilya Sutskever, and Geoffrey~E. Hinton.
\newblock Imagenet classification with deep convolutional neural networks.
\newblock In {\em {NeurIPS}}, pages 1106--1114, 2012.

\bibitem{NINH:conf/cvpr/LaiPLY15}
Hanjiang Lai, Yan Pan, Ye~Liu, and Shuicheng Yan.
\newblock Simultaneous feature learning and hash coding with deep neural
  networks.
\newblock In {\em CVPR}, pages 3270--3278, 2015.

\bibitem{DBLP:conf/ijcai/LiLDLG18}
Ning Li, Chao Li, Cheng Deng, Xianglong Liu, and Xinbo Gao.
\newblock Deep joint semantic-embedding hashing.
\newblock In {\em {IJCAI}}, pages 2397--2403, 2018.

\bibitem{DBLP:conf/nips/LiSHT17}
Qi~Li, Zhenan Sun, Ran He, and Tieniu Tan.
\newblock Deep supervised discrete hashing.
\newblock In {\em {NeurIPS}}, pages 2482--2491, 2017.

\bibitem{DBLP:conf/ijcai/LiWK16}
Wu{-}Jun Li, Sheng Wang, and Wang{-}Cheng Kang.
\newblock Feature learning based deep supervised hashing with pairwise labels.
\newblock In {\em {IJCAI}}, pages 1711--1717, 2016.

\bibitem{DBLP:journals/datamine/LiYZ23}
Xin{-}Chun Li, Yang Yang, and De{-}Chuan Zhan.
\newblock Mrtf: model refinery for transductive federated learning.
\newblock {\em DMKD}, 37(5):2046--2069, 2023.

\bibitem{DBLP:conf/eccv/LinMBHPRDZ14}
Tsung{-}Yi Lin, Michael Maire, Serge~J. Belongie, James Hays, Pietro Perona,
  Deva Ramanan, Piotr Doll{\'{a}}r, and C.~Lawrence Zitnick.
\newblock Microsoft {COCO:} common objects in context.
\newblock In {\em {ECCV}}, pages 740--755, 2014.

\bibitem{DBLP:conf/mm/LiuLZWHJ18}
Hong Liu, Mingbao Lin, Shengchuan Zhang, Yongjian Wu, Feiyue Huang, and
  Rongrong Ji.
\newblock Dense auto-encoder hashing for robust cross-modality retrieval.
\newblock In {\em {MM}}, pages 1589--1597, 2018.

\bibitem{DBLP:conf/nips/LiuMKC14}
Wei Liu, Cun Mu, Sanjiv Kumar, and Shih{-}Fu Chang.
\newblock Discrete graph hashing.
\newblock In {\em {NeurIPS}}, pages 3419--3427, 2014.

\bibitem{DBLP:conf/cvpr/LiuWJJC12}
Wei Liu, Jun Wang, Rongrong Ji, Yu{-}Gang Jiang, and Shih{-}Fu Chang.
\newblock Supervised hashing with kernels.
\newblock In {\em {CVPR}}, pages 2074--2081, 2012.

\bibitem{DBLP:conf/icml/LiuWKC11}
Wei Liu, Jun Wang, Sanjiv Kumar, and Shih{-}Fu Chang.
\newblock Hashing with graphs.
\newblock In {\em {ICML}}, pages 1--8, 2011.

\bibitem{DBLP:conf/mm/LiuNZCZY18}
Xingbo Liu, Xiushan Nie, Wenjun Zeng, Chaoran Cui, Lei Zhu, and Yilong Yin.
\newblock Fast discrete cross-modal hashing with regressing from semantic
  labels.
\newblock In {\em {MM}}, pages 1662--1669, 2018.

\bibitem{DBLP:conf/mm/LiuNZY19}
Xingbo Liu, Xiushan Nie, Quan Zhou, and Yilong Yin.
\newblock Supervised discrete hashing with mutual linear regression.
\newblock In {\em {MM}}, pages 1561--1568, 2019.

\bibitem{DBLP:conf/mm/LuZCLNZ19}
Xu~Lu, Lei Zhu, Zhiyong Cheng, Jingjing Li, Xiushan Nie, and Huaxiang Zhang.
\newblock Flexible online multi-modal hashing for large-scale multimedia
  retrieval.
\newblock In {\em {MM}}, pages 1129--1137, 2019.

\bibitem{DBLP:conf/mm/MaoWZW18}
Zhendong Mao, Quan Wang, Yongdong Zhang, and Bin Wang.
\newblock Post tuned hashing: {A} new approach to indexing high-dimensional
  data.
\newblock In {\em {MM}}, pages 834--842, 2018.

\bibitem{DBLP:journals/tgrs/MengWMYX24}
Lingwu Meng, Jing Wang, Ran Meng, Yang Yang, and Liang Xiao.
\newblock A multiscale grouping transformer with {CLIP} latents for remote
  sensing image captioning.
\newblock {\em TGRS}, 62:1--15, 2024.

\bibitem{DBLP:journals/nature/MnihKSRVBGRFOPB15}
Volodymyr Mnih, Koray Kavukcuoglu, David Silver, Andrei~A. Rusu, Joel Veness,
  Marc~G. Bellemare, Alex Graves, Martin~A. Riedmiller, Andreas Fidjeland,
  Georg Ostrovski, Stig Petersen, Charles Beattie, Amir Sadik, Ioannis
  Antonoglou, Helen King, Dharshan Kumaran, Daan Wierstra, Shane Legg, and
  Demis Hassabis.
\newblock Human-level control through deep reinforcement learning.
\newblock {\em Nature}, 518(7540):529--533, 2015.

\bibitem{DBLP:conf/icml/NorouziF11}
Mohammad Norouzi and David~J. Fleet.
\newblock Minimal loss hashing for compact binary codes.
\newblock In {\em {ICML}}, pages 353--360, 2011.

\bibitem{DBLP:conf/nips/PaszkeGMLBCKLGA19}
Adam Paszke, Sam Gross, Francisco Massa, Adam Lerer, James Bradbury, Gregory
  Chanan, Trevor Killeen, Zeming Lin, Natalia Gimelshein, Luca Antiga, Alban
  Desmaison, Andreas K{\"{o}}pf, Edward Yang, Zachary DeVito, Martin Raison,
  Alykhan Tejani, Sasank Chilamkurthy, Benoit Steiner, Lu~Fang, Junjie Bai, and
  Soumith Chintala.
\newblock Pytorch: An imperative style, high-performance deep learning library.
\newblock In {\em {NeurIPS}}, pages 8024--8035, 2019.

\bibitem{DBLP:conf/iccv/QiZC019}
Guo{-}Jun Qi, Liheng Zhang, Chang~Wen Chen, and Qi~Tian.
\newblock {AVT:} unsupervised learning of transformation equivariant
  representations by autoencoding variational transformations.
\newblock In {\em {ICCV}}, pages 8129--8138, 2019.

\bibitem{DBLP:conf/mm/RagnarsdottirK19}
Hanna Ragnarsd{\'{o}}ttir, {\TH}{\'{o}}rhildur {\TH}orleiksd{\'{o}}ttir,
  Omar~Shahbaz Khan, Bj{\"{o}}rn~{\TH}{\'{o}}r J{\'{o}}nsson,
  Gylfi~{\TH}{\'{o}}r Gu{\dh}mundsson, Jan Zah{\'{a}}lka, Stevan Rudinac,
  Laurent Amsaleg, and Marcel Worring.
\newblock Exquisitor: Breaking the interaction barrier for exploration of 100
  million images.
\newblock In {\em {MM}}, pages 1029--1031, 2019.

\bibitem{DBLP:conf/icdm/RazzakYYX19}
Farid Razzak, Fei Yi, Yang Yang, and Hui Xiong.
\newblock An integrated multimodal attention-based approach for bank stress
  test prediction.
\newblock In {\em ICDM}, pages 1282--1287. {IEEE}, 2019.

\bibitem{DBLP:conf/cvpr/ShenSLS15}
Fumin Shen, Chunhua Shen, Wei Liu, and Heng~Tao Shen.
\newblock Supervised discrete hashing.
\newblock In {\em {CVPR}}, pages 37--45, 2015.

\bibitem{DBLP:conf/cvpr/ShenSSHT13}
Fumin Shen, Chunhua Shen, Qinfeng Shi, Anton van~den Hengel, and Zhenmin Tang.
\newblock Inductive hashing on manifolds.
\newblock In {\em {CVPR}}, pages 1562--1569, 2013.

\bibitem{DBLP:journals/nature/SilverHMGSDSAPL16}
David Silver, Aja Huang, Chris~J. Maddison, Arthur Guez, Laurent Sifre, George
  van~den Driessche, Julian Schrittwieser, Ioannis Antonoglou, Vedavyas
  Panneershelvam, Marc Lanctot, Sander Dieleman, Dominik Grewe, John Nham, Nal
  Kalchbrenner, Ilya Sutskever, Timothy~P. Lillicrap, Madeleine Leach, Koray
  Kavukcuoglu, Thore Graepel, and Demis Hassabis.
\newblock Mastering the game of go with deep neural networks and tree search.
\newblock {\em Nature}, 529(7587):484--489, 2016.

\bibitem{DBLP:conf/nips/SuZHT18}
Shupeng Su, Chao Zhang, Kai Han, and Yonghong Tian.
\newblock Greedy hash: Towards fast optimization for accurate hash coding in
  {CNN}.
\newblock In {\em NeurIPS}, pages 806--815, 2018.

\bibitem{DBLP:journals/pami/TangSQLWYJ17}
Jinhui Tang, Xiangbo Shu, Guo{-}Jun Qi, Zechao Li, Meng Wang, Shuicheng Yan,
  and Ramesh~C. Jain.
\newblock Tri-clustered tensor completion for social-aware image tag
  refinement.
\newblock {\em {TPAMI}}, 39(8):1662--1674, 2017.

\bibitem{DBLP:conf/cvpr/TomeiCBC19}
Matteo Tomei, Marcella Cornia, Lorenzo Baraldi, and Rita Cucchiara.
\newblock Art2real: Unfolding the reality of artworks via semantically-aware
  image-to-image translation.
\newblock In {\em {CVPR}}, pages 5849--5859, 2019.

\bibitem{DBLP:conf/icme/WanWGY24}
Fengqiang Wan, WU~Xiangyu, Zhihao Guan, and Yang Yang.
\newblock Covlr: Coordinating cross-modal consistency and intra-modal relations
  for vision-language retrieval.
\newblock ICME, 2024.

\bibitem{DBLP:journals/pami/WangZSSS18}
Jingdong Wang, Ting Zhang, Jingkuan Song, Nicu Sebe, and Heng~Tao Shen.
\newblock A survey on learning to hash.
\newblock {\em {TPAMI}}, 40(4):769--790, 2018.

\bibitem{RSH:conf/iccv/WangLSJ13}
Jun Wang, Wei Liu, Andy~X. Sun, and Yu{-}Gang Jiang.
\newblock Learning hash codes with listwise supervision.
\newblock In {\em {ICCV}}, pages 3032--3039, 2013.

\bibitem{DBLP:journals/ml/Williams92}
Ronald~J. Williams.
\newblock Simple statistical gradient-following algorithms for connectionist
  reinforcement learning.
\newblock {\em ML}, 8:229--256, 1992.

\bibitem{DBLP:conf/aaai/XiaPLLY14}
Rongkai Xia, Yan Pan, Hanjiang Lai, Cong Liu, and Shuicheng Yan.
\newblock Supervised hashing for image retrieval via image representation
  learning.
\newblock In {\em {AAAI}}, pages 2156--2162, 2014.

\bibitem{DBLP:journals/pami/XuROWS19}
Dan Xu, Elisa Ricci, Wanli Ouyang, Xiaogang Wang, and Nicu Sebe.
\newblock Monocular depth estimation using multi-scale continuous crfs as
  sequential deep networks.
\newblock {\em {TPAMI}}, 41(6):1426--1440, 2019.

\bibitem{DBLP:conf/mm/YanP0S0H19}
Cheng Yan, Guansong Pang, Xiao Bai, Chunhua Shen, Jun Zhou, and Edwin~R.
  Hancock.
\newblock Deep hashing by discriminating hard examples.
\newblock In {\em {MM}}, pages 1535--1542, 2019.

\bibitem{DBLP:journals/chinaf/YangBGZYY23}
Yang Yang, Ran Bao, Weili Guo, De{-}Chuan Zhan, Yilong Yin, and Jian Yang.
\newblock Deep visual-linguistic fusion network considering cross-modal
  inconsistency for rumor detection.
\newblock {\em SCIS}, 66(12), 2023.

\bibitem{DBLP:journals/tkde/YangFZLJ21}
Yang Yang, Zhao{-}Yang Fu, De{-}Chuan Zhan, Zhi{-}Bin Liu, and Yuan Jiang.
\newblock Semi-supervised multi-modal multi-instance multi-label deep network
  with optimal transport.
\newblock {\em TKDE}, 33(2):696--709, 2021.

\bibitem{DBLP:journals/fcsc/YangGLLLY24}
Yang Yang, Jinyi Guo, Guangyu Li, Lanyu Li, Wenjie Li, and Jian Yang.
\newblock Alignment efficient image-sentence retrieval considering transferable
  cross-modal representation learning.
\newblock {\em FCS}, 18(3):181335, 2024.

\bibitem{DBLP:conf/aaai/YangHGXX23}
Yang Yang, Yurui Huang, Weili Guo, Baohua Xu, and Dingyin Xia.
\newblock Towards global video scene segmentation with context-aware
  transformer.
\newblock In {\em AAAI}, pages 3206--3213. {AAAI} Press, 2023.

\bibitem{DBLP:journals/tkdd/YangWSLZXY22}
Yang Yang, Hongchen Wei, Zhen{-}Qiang Sun, Guangyu Li, Yuanchun Zhou, Hui
  Xiong, and Jian Yang.
\newblock {S2OSC:} {A} holistic semi-supervised approach for open set
  classification.
\newblock {\em TKDD}, 16(2):34:1--34:27, 2022.

\bibitem{DBLP:journals/toc/YangWZYXY22}
Yang Yang, Hongchen Wei, Hengshu Zhu, Dianhai Yu, Hui Xiong, and Jian Yang.
\newblock Exploiting cross-modal prediction and relation consistency for
  semisupervised image captioning.
\newblock {\em TOC}, 2022.

\bibitem{DBLP:journals/tkde/YangYBZZGXY23}
Yang Yang, Jia{-}Qi Yang, Ran Bao, De{-}Chuan Zhan, Hengshu Zhu, Xiaoru Gao,
  Hui Xiong, and Jian Yang.
\newblock Corporate relative valuation using heterogeneous multi-modal graph
  neural network.
\newblock {\em TKDE}, 35(1):211--224, 2023.

\bibitem{DBLP:journals/tkde/YangZWLXJ21}
Yang Yang, De{-}Chuan Zhan, Yi{-}Feng Wu, Zhi{-}Bin Liu, Hui Xiong, and Yuan
  Jiang.
\newblock Semi-supervised multi-modal clustering and classification with
  incomplete modalities.
\newblock {\em TKDE}, 33(2):682--695, 2021.

\bibitem{DBLP:journals/tois/YangZSDZL24}
Yang Yang, Chubing Zhang, Xin Song, Zheng Dong, Hengshu Zhu, and Wenjie Li.
\newblock Contextualized knowledge graph embedding for explainable talent
  training course recommendation.
\newblock {\em TOIS}, 42(2):33:1--33:27, 2024.

\bibitem{DBLP:conf/ijcai/YangZXYZY21}
Yang Yang, Chubing Zhang, Yi{-}Chu Xu, Dianhai Yu, De{-}Chuan Zhan, and Jian
  Yang.
\newblock Rethinking label-wise cross-modal retrieval from {A} semantic sharing
  perspective.
\newblock In {\em IJCAI}, pages 3300--3306. ijcai.org, 2021.

\bibitem{DBLP:conf/mm/0074ZGGZ22}
Yang Yang, Jingshuai Zhang, Fan Gao, Xiaoru Gao, and Hengshu Zhu.
\newblock {DOMFN:} {A} divergence-orientated multi-modal fusion network for
  resume assessment.
\newblock In {\em MM}, pages 1612--1620. {ACM}, 2022.

\bibitem{DBLP:conf/kdd/YangZZX019}
Yang Yang, Da{-}Wei Zhou, De{-}Chuan Zhan, Hui Xiong, and Yuan Jiang.
\newblock Adaptive deep models for incremental learning: Considering capacity
  scalability and sustainability.
\newblock In {\em KDD}, pages 74--82. {ACM}, 2019.

\bibitem{DBLP:conf/mm/YeP18}
Zhaoda Ye and Yuxin Peng.
\newblock Multi-scale correlation for sequential cross-modal hashing learning.
\newblock In {\em {MM}}, pages 852--860, 2018.

\bibitem{DBLP:journals/corr/abs-1802-02904}
Jian Zhang, Yuxin Peng, and Zhaoda Ye.
\newblock Deep reinforcement learning for image hashing.
\newblock {\em CoRR}, abs/1802.02904, 2018.

\bibitem{DRSH:journals/tip/ZhangLZZZ15}
Ruimao Zhang, Liang Lin, Rui Zhang, Wangmeng Zuo, and Lei Zhang.
\newblock Bit-scalable deep hashing with regularized similarity learning for
  image retrieval and person re-identification.
\newblock {\em {TIP}}, 24(12):4766--4779, 2015.

\bibitem{DBLP:journals/titb/ZhangYYGLZYR22}
Xiao Zhang, Hongzheng Yu, Yang Yang, Jingjing Gu, Yujun Li, Fuzhen Zhuang,
  Dongxiao Yu, and Zhaochun Ren.
\newblock Harmi: Human activity recognition via multi-modality incremental
  learning.
\newblock {\em TITB}, 26(3):939--951, 2022.

\end{thebibliography}

\end{document}